# STC-MGNet: Spatial-Temporal-Channel Attention-Enhanced Multimodal Graph Network for Pedestrian Intention Prediction

Yuanzhe Li
Chair of Automotive Engineering
Technische Universität Berlin
Berlin, Germany

Meng Wang
Chair of Automotive Engineering
Technische Universität Berlin
Berlin, Germany

***Abstract*—Accurate prediction of pedestrian crossing intentions plays a critical role in enhancing the safety of pedestrian-vehicle interactions in autonomous driving. However, effectively capturing subtle spatiotemporal patterns in pedestrian pose sequences and integrating complementary visual and ego-vehicle motion information remain challenging. This paper proposes a spatial-temporal-channel attention-enhanced multimodal graph network (STC-MGNet) for pedestrian crossing intention prediction. STC-MGNet introduces a spatial-temporal-channel (STC) attention-enhanced pose graph encoder to learn discriminative representations from pedestrian pose sequences. Three parallel attention branches selectively emphasize informative skeletal joints, observation frames, and feature channels, with their outputs integrated through learnable fusion weights. To leverage complementary visual and ego-vehicle motion information, a multi-stage cross-modal channel modulation mechanism is introduced to adaptively refine pose representations through modality-specific channel gating at successive graph encoding stages. Experimental results demonstrate that STC-MGNet achieves accuracy rates of 70%, 87%, and 88% on the $JAAD_{beh}$, $JAAD_{all}$, and PIE datasets, respectively, outperforming several representative baseline methods. Ablation experiments further highlight the complementary contributions of visual and ego-vehicle motion information to prediction performance.**

## I. INTRODUCTION

Recent advances in computing and artificial intelligence have accelerated the development and deployment of autonomous vehicles (AVs) in urban environments [1]. As a result, concerns regarding the safety of vulnerable road users (VRUs), particularly pedestrians, during interactions with AVs have received increasing attention [2]. Unlike vehicle occupants, pedestrians lack the protection provided by vehicle structures and are consequently more susceptible to severe or fatal injuries in traffic collisions. Predicting pedestrian crossing intention enables AVs to anticipate potential crossing behaviors and supports timely driving decisions and vehicle control. Nevertheless, reliable prediction of pedestrian crossing intentions is complicated by diverse behavioral patterns, subtle postural changes, and dynamic traffic conditions.

Pedestrian pose sequences capture the spatial configurations of skeletal joints and their temporal dynamics, providing discriminative behavioral information for crossing intention prediction. Early studies employed handcrafted pose features and conventional machine learning classifiers, such as support vector machines (SVMs) and random forests (RFs), to predict pedestrian crossing intentions [3, 4]. Subsequent studies have adopted graph convolutional networks (GCNs) to characterize spatial relationships among skeletal joints [5]. Attention mechanisms have further been incorporated into spatiotemporal graph architectures to enhance the extraction of discriminative skeletal features and temporal dynamics [7]. Despite these advances, relying solely on pose information may be insufficient to distinguish crossing intentions when pedestrian movements are subtle or ambiguous. Furthermore, pose keypoints alone cannot capture complementary visual cues or ego-vehicle motion information, both of which can provide valuable information for pedestrian crossing intention prediction.

Motivated by these limitations, subsequent studies have investigated multimodal architectures that leverage complementary visual and motion cues for pedestrian crossing intention prediction. For example, PCPA [8] employs temporal and modality attention mechanisms to fuse visual, pose, and motion features, while Action-ViT [9] progressively integrates multimodal features using a Vision Transformer-based architecture. More recent studies have further incorporated depth maps [10], optical flow [11], and high-level semantic contextual attributes [12] to enrich geometric perception, motion characterization, and contextual understanding. Despite these advances, existing multimodal fusion approaches primarily focus on integrating modality-specific features into joint representations, which does not necessarily ensure the effective incorporation of complementary visual and ego-vehicle motion information into skeletal feature learning. Therefore, effectively leveraging such information to refine pose representations remains an important challenge.

In this paper, we propose STC-MGNet, a spatial-temporal-channel attention-enhanced multimodal graph network for pedestrian crossing intention prediction. The proposed network employs an STC-enhanced pose graph encoder to model spatial dependencies among skeletal joints and temporal dependencies across observation frames. Three parallel attention branches further refine pose representations along the spatial, temporal, and channel dimensions, with their outputs integrated through learnable fusion weights. Furthermore, a multi-stage cross-modal channel modulation mechanism is introduced to refine pose representations using visual and ego-vehicle motion information through modality-specific gating. Finally, the refined pose representations are aggregated and fed into a multilayer perceptron (MLP) for intention prediction.

The main contributions are summarized as follows:

1. We develop an STC-enhanced pose graph encoder that incorporates three parallel attention branches along the spatial, temporal, and channel dimensions, with their outputs integrated through learnable fusion weights to enhance discriminative pose representations for pedestrian crossing intention prediction.

2. We design a multi-stage cross-modal channel modulation mechanism that leverages complementary visual and ego-vehicle motion information to progressively refine pose representations through modality-specific channel gating.

3. Extensive experiments on the JAAD and PIE datasets demonstrate the competitive performance of the proposed STC-MGNet against state-of-the-art methods. Ablation studies further evaluate the individual and complementary contributions of different input modalities.

## II. Methodology

### A. Overall Architecture

The overall architecture of the proposed STC-MGNet is illustrated in Fig. 1. The network takes three input modalities: pedestrian pose keypoints, local RGB images, and ego-vehicle speed. An STC-enhanced pose graph encoder is employed to capture spatial dependencies among skeletal joints and temporal dependencies across observation frames. Meanwhile, local RGB images and ego-vehicle speed are processed using 2D and 1D convolutional layers, respectively, to extract complementary visual and ego-vehicle motion features. The extracted visual and ego-vehicle motion features are subsequently used to refine pose representations through modality-specific channel gating at different graph encoding stages. Finally, the fused features are aggregated through global average pooling and fed into an MLP for pedestrian crossing intention prediction.

### B. STC-Enhanced Pose Graph Encoding

The STC-enhanced pose graph encoder is designed to capture spatial dependencies among skeletal joints and temporal dependencies across observation frames while refining pose representations along the spatial, temporal, and channel dimensions. Given a pedestrian pose sequence $X \in \mathbb{R}^{B \times C \times T \times N}$, where $B$, $C$, $T$, and $N$ denote the batch size, feature channels, sequence length, and number of skeletal joints, respectively. The human skeleton is represented as a graph $\mathcal{G} = (\mathcal{V}, \mathcal{E})$, with skeletal joints defined as nodes and anatomical connections represented as edges. Following the spatial partitioning strategy of ST-GCN [21], the graph connections are divided into three subsets: self-links, inward connections, and outward connections. The graph convolution operation is formulated as follows:

$$\mathbf{H}_g = \sigma\left(\sum_{k=1}^{3} \tilde{\mathbf{A}}^{(k)} \mathbf{X} \mathbf{W}^{(k)}\right) \tag{1}$$

where $\tilde{\mathbf{A}}^{(k)}$ denotes the normalized adjacency matrix of the $k$-th subset, $\mathbf{W}^{(k)}$ represents the corresponding learnable transformation matrix, and $\sigma(\cdot)$ denotes the activation function.

The graph convolutional layer learns structural pose representations through feature aggregation across anatomically connected skeletal joints. Within each STC-enhanced pose graph encoding block, an STC attention module is applied to the graph convolution output, as illustrated in Fig. 2. The module comprises three parallel spatial, temporal, and channel attention branches that refine pose features along their respective dimensions.

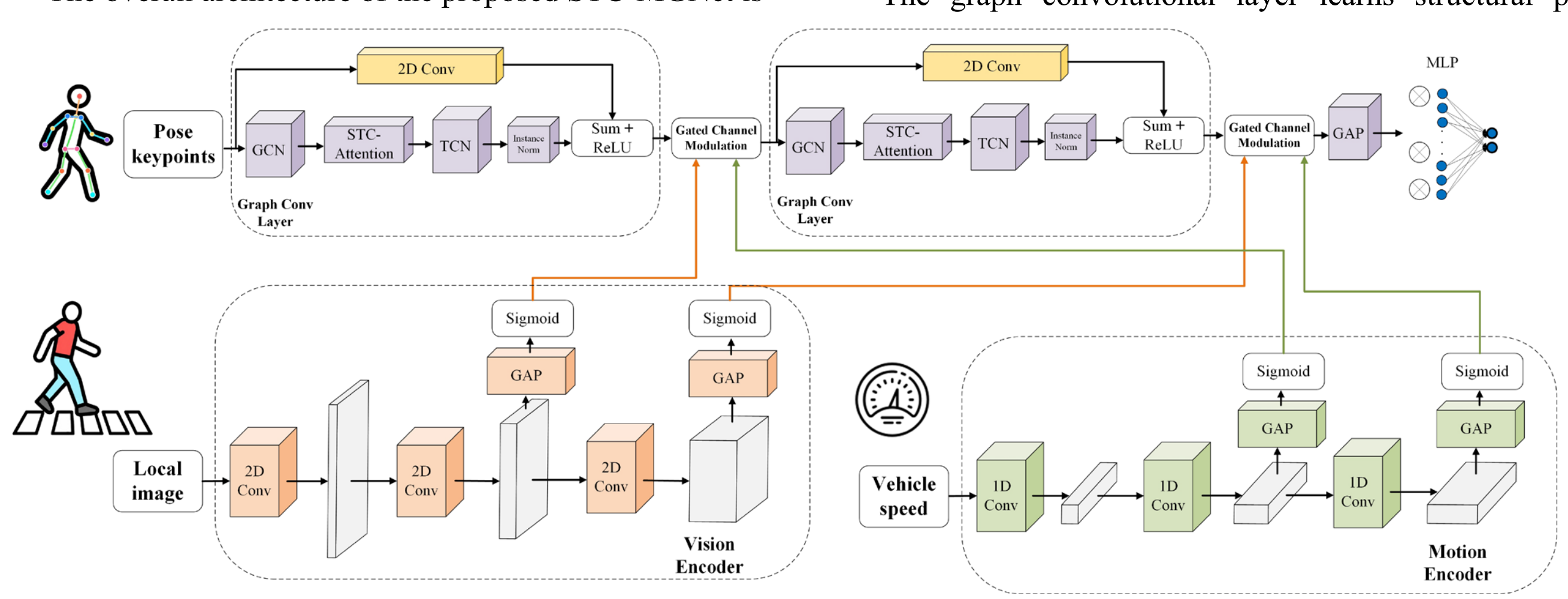

Figure 1. Overall architecture of the proposed STC-MGNet.

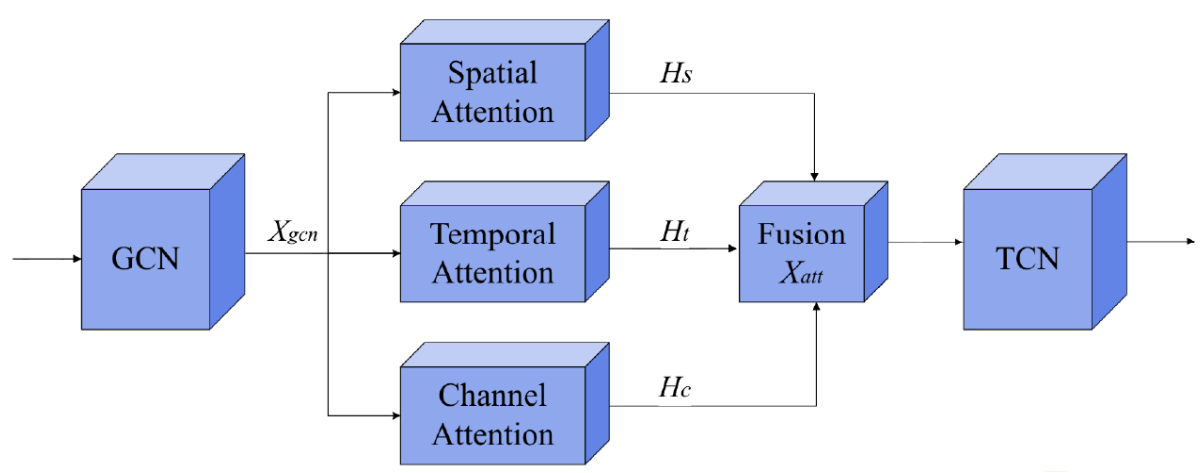

Figure 2. Architecture of the STC-enhanced graph encoding block.

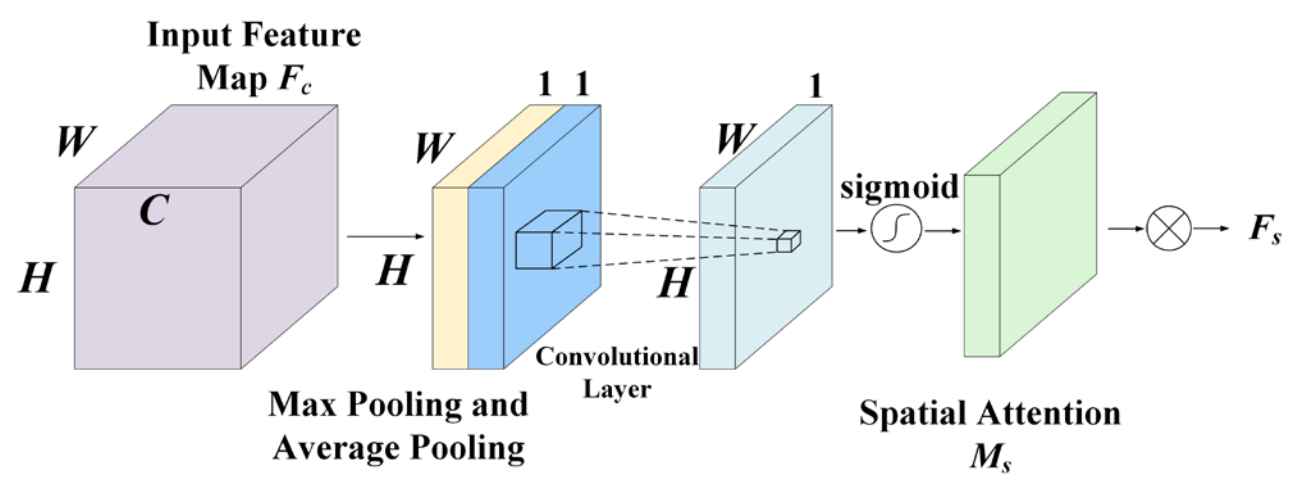

Figure 3. Architecture of the spatial attention branch.

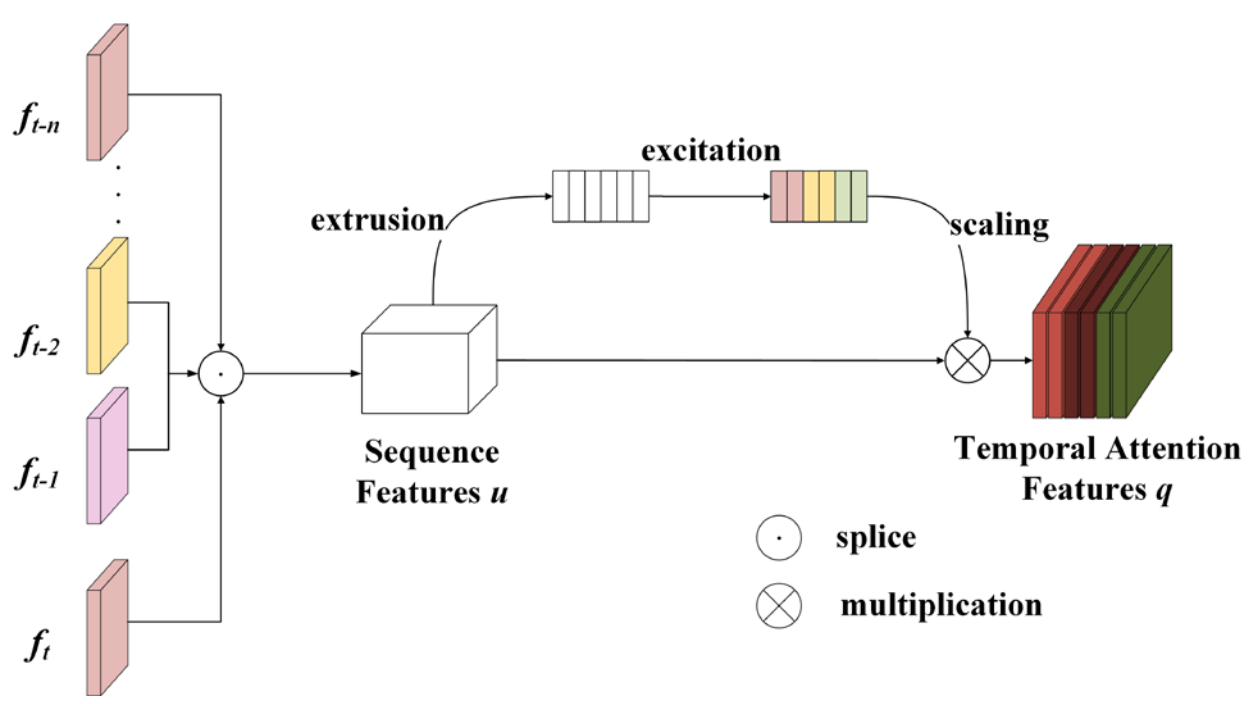

Figure 4. Architecture of the temporal attention branch.

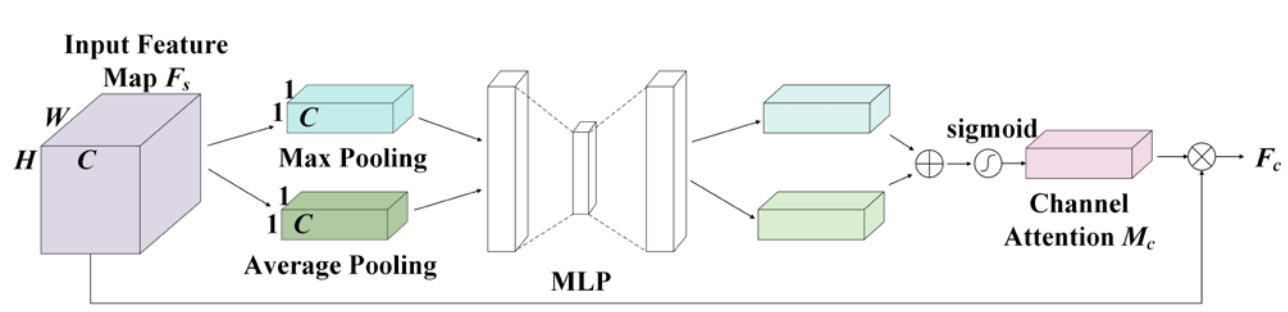

Figure 5. Architecture of the channel attention branch.

Specifically, the spatial attention branch assigns joint-wise importance weights to the pose representations, as illustrated in Fig. 3. The spatial attention map is computed as

$$\mathbf{A}_s = \mathrm{Sigmoid}\left(\mathrm{Conv}_{1\times 1}\left(\mathrm{AvgPool}_T(\mathbf{H}_g)\right)\right) \tag{2}$$

where $\mathrm{AvgPool}_T(\cdot)$ denotes average pooling along the temporal dimension and $\mathrm{Conv}_{1\times 1}(\cdot)$ represents a $1\times 1$ convolutional operation. The resulting attention map is used to modulate the graph features along the spatial dimension.

Similarly, the temporal attention branch, illustrated in Fig. 4, generates a temporal attention map to refine pose features across observation frames. The temporal attention map is formulated as follows:

$$\mathbf{A}_t = \mathrm{Sigmoid}\left(f_t\left(\mathrm{AvgPool}_N(\mathbf{H}_g)\right)\right) \tag{3}$$

where $\mathrm{AvgPool}_N(\cdot)$ denotes average pooling over the skeletal joints, and $f_t(\cdot)$ represents a 1D convolutional operation for temporal attention generation.

In parallel, the channel attention branch, illustrated in Fig. 5, models interdependencies among feature channels to refine channel-wise feature representations. The channel attention map is obtained as

$$\mathbf{A}_c = \mathrm{Sigmoid}\left(\mathbf{W}_{c1}\,\mathrm{ReLU}\left(\mathbf{W}_{c2}\,\mathrm{GAP}(\mathbf{H}_g)\right)\right) \tag{4}$$

where $\mathrm{GAP}(\cdot)$ denotes global average pooling over the temporal and joint dimensions, $\mathbf{W}_{c1}$ and $\mathbf{W}_{c2}$ are learnable projection matrices, and $\delta(\cdot)$ represents the ReLU activation function.

The attention maps generated by the three branches are applied to the GCN output through element-wise multiplication, yielding the corresponding attention-enhanced pose representations:

$$\mathbf{H}_s = \mathbf{H}_g \odot \mathbf{A}_s \tag{5}$$

$$\mathbf{H}_t = \mathbf{H}_g \odot \mathbf{A}_t \tag{6}$$

$$\mathbf{H}_c = \mathbf{H}_g \odot \mathbf{A}_c \tag{7}$$

where $\odot$ denotes element-wise multiplication.

As illustrated in Fig. 2, the three attention-enhanced representations are integrated through a learnable weighted fusion mechanism. Specifically, three trainable scalar parameters $\alpha_s$, $\alpha_t$, and $\alpha_c$ are introduced to control the contributions of the spatial, temporal, and channel attention branches, respectively. These parameters are normalized using a softmax function:

$$[w_s, w_t, w_c] = \mathrm{Softmax}([\alpha_s, \alpha_t, \alpha_c]) \tag{8}$$

where $w_s$, $w_t$, and $w_c$ denote the corresponding normalized fusion weights. The resulting STC-enhanced pose representation is formulated as:

$$\mathbf{H}_{att} = w_s\mathbf{H}_s + w_t\mathbf{H}_t + w_c\mathbf{H}_c \tag{9}$$

Subsequently, a temporal convolutional network (TCN) is applied to the fused pose representation to capture temporal dependencies across observation frames. Instance normalization is performed on the TCN output, followed by residual addition and ReLU activation. The output of the STC-enhanced pose graph encoding block is formulated as follows:

$$\mathbf{Y} = \mathrm{ReLU}\left(\mathrm{IN}\left(\mathrm{TCN}(\mathbf{H}_{att})\right) + \mathbf{X}_{res}\right) \tag{10}$$

where IN(·) denotes instance normalization and and $\mathbf{X}_{res}$ represents the residual mapping of the block input. The TCN employs a convolutional kernel of size (5, 1) to model temporal dependencies while preserving the skeletal joint dimension.

### C. Visual and Ego-Vehicle Motion Encoding

To complement the structural pose representations, two lightweight CNN encoders are employed to extract visual and ego-vehicle motion features. The visual encoder consists of three sequential 2D convolutional layers to extract features from local RGB images, while the ego-vehicle motion encoder employs three sequential 1D convolutional layers to capture temporal variations in ego-vehicle speed. For each encoder, global average pooling (GAP) is independently applied to the feature maps from the second and third convolutional layers, yielding channel-wise visual and ego-vehicle motion feature descriptors. These descriptors are subsequently used to generate modality-specific channel gating weights for refining pose representations through cross-modal channel modulation.

### D. Multi-Stage Cross-Modal Channel Modulation

To integrate complementary visual and ego-vehicle motion information, a multi-stage cross-modal channel modulation mechanism is employed to refine the structural pose representations at different graph encoding stages. Specifically, the visual and ego-vehicle motion feature descriptors are transformed into modality-specific gating weights to regulate their contributions to the structural pose representations. At the $l$-th modulation stage, let $\mathbf{z}_I^{(l)}$ and $\mathbf{z}_V^{(l)}$ denote the channel-wise visual and motion feature descriptors, respectively. The gating weights are generated through a sigmoid activation function:

$$\mathbf{g}_I^{(l)} = \sigma\left(\mathbf{z}_I^{(l)}\right) \tag{11}$$

$$\mathbf{g}_V^{(l)} = \sigma\left(\mathbf{z}_V^{(l)}\right) \tag{12}$$

where $\mathbf{g}_I^{(l)}$ and $\mathbf{g}_V^{(l)}$ represent the modality-specific channel gating weights for visual and ego-vehicle motion features, respectively. These weights are subsequently employed to modulate the structural pose representations through a residual gating mechanism, formulated as

$$\mathbf{H}_{\text{fused}}^{(l)} = \mathbf{H}_p^{(l)} \odot \left(1 + \lambda_I^{(l)}\mathbf{g}_I^{(l)} + \lambda_V^{(l)}\mathbf{g}_V^{(l)}\right) \tag{13}$$

where $\mathbf{H}_p^{(l)}$ denotes the pose representation extracted at the $l$-th graph encoding stage, and $\lambda_I^{(l)}$ and $\lambda_V^{(l)}$ are learnable scaling parameters that control the respective contributions of visual and ego-vehicle motion information. The gating weights are broadcast across the temporal and skeletal joint dimensions for channel-wise modulation, while the identity term preserves a direct contribution from the original pose representation.

By applying the gated modulation mechanism at multiple graph encoding stages, complementary visual and ego-vehicle motion information is progressively integrated into hierarchical pose representations. This process enables cross-modal feature refinement while preserving the spatial and temporal structure of the pose representations.

## III. Experiments

### A. Datasets and Implementation Details

STC-MGNet is evaluated on the JAAD [22] and PIE [23] datasets, both comprising on-board videos recorded at 30 FPS. JAAD comprises two subsets, $\mathrm{JAAD_{beh}}$ and $\mathrm{JAAD_{all}}$, and follows a benchmark split of 177 training, 117 testing, and 29 validation videos. The PIE dataset is divided into 50% for training, 40% for testing, and 10% for validation. For both datasets, 32-frame observation sequences (approximately 1 s) are extracted for each target pedestrian using a sliding window with a temporal stride of one frame. The pedestrian pose in each observed frame is represented by 19 skeletal keypoints, while visual information is extracted from an image region twice the size of the pedestrian bounding box. Ego-vehicle motion information is obtained from dataset annotations, comprising discrete vehicle-action states in JAAD and continuous speed measurements in PIE.

STC-MGNet is implemented in PyTorch and trained on an NVIDIA GeForce RTX 4080 Laptop GPU. For both JAAD and PIE, the graph backbone consists of two STC-enhanced graph encoding blocks. The models are trained on JAAD and PIE for 30 epochs using the AdamW optimizer, with a batch size of 32 and a weight decay of $1\times10^{-3}$. The initial learning rates are set to $5\times10^{-3}$ and $2\times10^{-4}$ for JAAD and PIE, respectively. A ReduceLROnPlateau scheduler is used for learning rate adjustment.

### B. Comparison with State-of-the-Art Methods

To evaluate the effectiveness of the proposed method, we compare its performance with representative state-of-the-art (SOTA) methods on the JAAD and PIE datasets. The quantitative results are presented in Table I. Performance is evaluated using five metrics: accuracy (Acc), area under the ROC curve (AUC), F1-score (F1), precision (P), and recall (R).

On the $\mathrm{JAAD_{beh}}$ dataset, STC-MGNet achieves the highest accuracy, AUC, F1 score, and precision, reaching 0.70, 0.70, 0.80, and 0.70, respectively. Compared with RU-LSTM [13], STC-MGNet improves accuracy by 0.01, AUC by 0.08, and F1 score by 0.02. Moreover, the proposed method achieves a recall of 0.92, ranking second behind LSOP-Net [16] (0.98). On the $\mathrm{JAAD_{all}}$ dataset, STC-MGNet achieves the highest accuracy of 0.87, outperforming the second-best methods by 0.01. It also achieves a precision of

TABLE I. PERFORMANCE COMPARISON WITH SOTA METHODS ON JAAD AND PIE DATASETS

| **Models** | **$JAAD_{beh}$** | | | | | **$JAAD_{all}$** | | | | | **PIE** | | | | |
|---|---|---|---|---|---|---|---|---|---|---|---|---|---|---|---|
| | **Acc** | **AUC** | **F1** | **P** | **R** | **Acc** | **AUC** | **F1** | **P** | **R** | **Acc** | **AUC** | **F1** | **P** | **R** |
| SingleRNN [6] | 0.51 | 0.48 | 0.61 | 0.63 | 0.59 | 0.78 | 0.75 | 0.54 | 0.44 | 0.70 | 0.81 | 0.75 | 0.64 | 0.67 | 0.61 |
| MultiRNN [14] | 0.61 | 0.50 | 0.74 | 0.64 | 0.86 | 0.79 | 0.79 | 0.58 | 0.45 | 0.79 | 0.83 | 0.80 | 0.71 | 0.69 | 0.73 |
| SFRNN [24] | 0.51 | 0.45 | 0.63 | 0.61 | 0.64 | 0.84 | 0.84 | 0.65 | 0.54 | 0.84 | 0.82 | 0.79 | 0.69 | 0.67 | 0.70 |
| PCPA [8] | 0.56 | 0.54 | 0.63 | **0.66** | 0.60 | 0.77 | 0.79 | 0.56 | 0.42 | 0.83 | **0.87** | 0.85 | 0.78 | 0.76 | **0.81** |
| RU-LSTM [13] | **0.69** | **0.62** | **0.78** | - | - | **0.86** | 0.78 | 0.62 | - | - | **0.87** | 0.84 | 0.77 | - | - |
| Action-ViT [9] | - | - | - | - | - | **0.86** | **0.88** | **0.90** | **0.85** | **0.96** | 0.86 | **0.89** | 0.78 | **0.87** | 0.71 |
| ST CrossingPose [15] | 0.63 | 0.56 | 0.74 | **0.66** | 0.83 | - | - | - | - | - | - | - | - | - | - |
| LSOP-Net [16] | 0.65 | 0.54 | **0.78** | 0.65 | **0.98** | **0.85** | 0.75 | 0.58 | 0.56 | 0.61 | **0.89** | 0.87 | **0.81** | **0.80** | **0.82** |
| PedGNN [17] | - | - | - | - | - | 0.80 | - | 0.85 | **0.85** | **0.88** | 0.68 | - | 0.70 | 0.67 | 0.68 |
| TrEP [18] | - | - | - | - | - | **0.86** | **0.94** | **0.86** | - | - | **0.87** | **0.93** | **0.87** | - | - |
| SAM [19] | **0.67** | **0.60** | **0.77** | **0.68** | **0.90** | 0.83 | 0.80 | 0.62 | 0.47 | 0.82 | - | - | - | - | - |
| GATEPose [20] | - | - | - | - | - | **0.85** | **0.93** | **0.87** | **0.86** | **0.89** | **0.87** | **0.90** | **0.86** | **0.83** | **0.88** |
| STC-MGNet | **0.70** | **0.70** | **0.80** | **0.70** | **0.92** | **0.87** | 0.85 | 0.52 | **0.74** | 0.41 | **0.88** | 0.88 | 0.78 | 0.78 | 0.77 |

Note: Green, blue, and orange indicate the best, second-best, and third-best results, respectively.

TABLE II. ABLATION STUDY OF MODALITY COMBINATIONS ON JAAD AND PIE DATASETS

| **Input Modalities** | **$JAAD_{beh}$** | | | | | **$JAAD_{all}$** | | | | | **PIE** | | | | |
|---|---|---|---|---|---|---|---|---|---|---|---|---|---|---|---|
| | **Acc** | **AUC** | **F1** | **P** | **R** | **Acc** | **AUC** | **F1** | **P** | **R** | **Acc** | **AUC** | **F1** | **P** | **R** |
| Pose | 0.55 | 0.49 | 0.25 | 0.18 | 0.41 | 0.75 | 0.69 | 0.31 | 0.38 | 0.25 | 0.77 | 0.77 | 0.66 | 0.57 | 0.67 |
| Pose + Image | 0.66 | 0.69 | 0.40 | 0.29 | 0.65 | 0.85 | **0.85** | 0.41 | **0.74** | 0.28 | 0.81 | 0.82 | 0.64 | 0.69 | 0.60 |
| Pose + Vehicle Speed | 0.60 | **0.71** | 0.39 | 0.27 | 0.72 | 0.77 | 0.69 | 0.42 | 0.38 | 0.25 | 0.84 | **0.88** | 0.69 | 0.77 | 0.62 |
| Pose + Image + Vehicle Speed | **0.70** | 0.70 | **0.80** | **0.70** | **0.92** | **0.87** | **0.85** | **0.52** | **0.74** | **0.41** | **0.88** | **0.88** | **0.78** | **0.78** | **0.77** |

0.74, ranking third among the compared methods. On the PIE dataset, STC-MGNet achieves an accuracy of 0.88, ranking second among the compared approaches. Compared with PCPA [8], it improves accuracy and AUC by 0.01 and 0.03, respectively, while achieving an identical F1 score of 0.78 and a higher precision of 0.78.

Overall, STC-MGNet achieves competitive performance against state-of-the-art methods on the JAAD and PIE datasets. These findings suggest that combining STC-enhanced pose graph encoding with multi-stage cross-modal channel modulation enables effective integration of complementary visual and ego-vehicle motion information to refine pose representations for pedestrian crossing intention prediction.

## C. *Ablation Studies*

To assess the individual and complementary contributions of different input modalities, four configurations are evaluated: (1) pedestrian pose alone, (2) pedestrian pose combined with local RGB images, (3) pedestrian pose combined with ego-vehicle speed, and (4) the complete STC-MGNet incorporating all three modalities. The results on the JAAD and PIE datasets are presented in Table II.

On $JAAD_{beh}$, the complete STC-MGNet achieves an accuracy of 0.70 and an F1 score of 0.80, substantially outperforming the pose-only configuration (0.55 and 0.25). Although incorporating ego-vehicle speed yields the highest AUC of 0.71, combining all three modalities achieves the highest recall of 0.92. On $JAAD_{all}$, visual information improves accuracy and precision to 0.85 and 0.74, respectively, while the complete model further increases accuracy and F1 score to 0.87 and 0.52. On PIE, the complete model achieves the highest accuracy of 0.88 and F1 score of 0.78, while matching the best AUC of 0.88.

Overall, the ablation results indicate that visual and ego-vehicle motion information provide complementary cues to pose representations, and their joint integration through the proposed multi-stage cross-modal channel modulation mechanism improves prediction performance.

## D. *Qualitative Analysis*

Fig. 6 presents three representative qualitative examples from the PIE dataset. The target pedestrian is highlighted by a red bounding box, with the predicted intention and corresponding ground truth provided for each case. In case (a), a pedestrian initially exhibits slight movement toward the roadway but subsequently stops near the curb. STC-MGNet correctly predicts No Crossing despite these subtle behavioral changes. In case (b), a pedestrian turns toward the roadway and initiates forward movement in a visually cluttered construction area. The network correctly predicts Crossing under these challenging visual conditions. In case (c), the target pedestrian approaches the roadway alongside another pedestrian in a complex traffic scene. Despite overl-apping motion patterns and surrounding traffic, STC-MGNet correctly predicts Crossing.

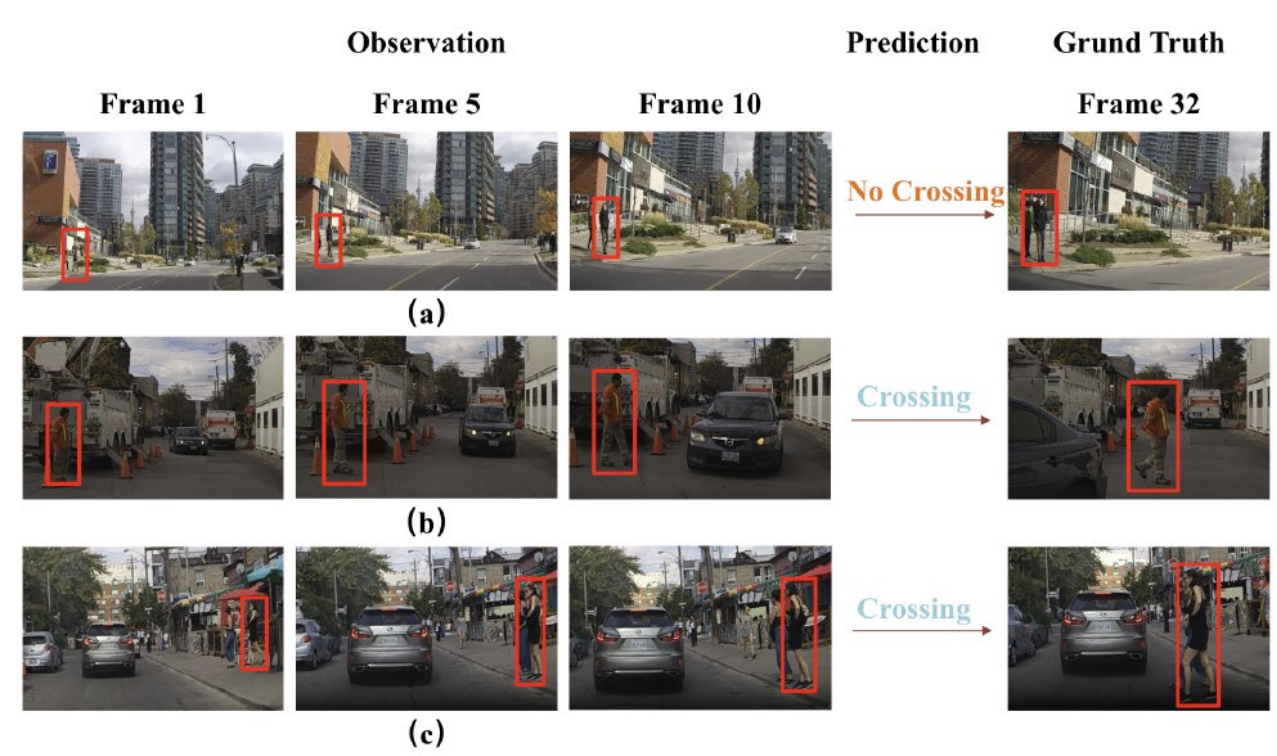

Figure 6. Qualitative results of pedestrian crossing intention prediction on the PIE dataset.

## IV. Conclusion

In this paper, we proposed STC-MGNet, a spatial-temporal-channel attention-enhanced multimodal graph network for pedestrian crossing intention prediction. The proposed framework employs an STC-enhanced pose graph encoder to model spatial dependencies among skeletal joints and temporal dynamics in pedestrian pose sequences. A multi-stage cross-modal channel modulation mechanism further integrates complementary visual and ego-vehicle motion information through modality-specific gating weights to refine hierarchical pose representations. Experimental evaluations on the JAAD and PIE datasets indicate that STC-MGNet achieves competitive performance compared with state-of-the-art methods, attaining accuracies of 70%, 87%, and 88% on $JAAD_{beh}$, $JAAD_{all}$, and PIE, respectively. Ablation studies further highlight the complementary contributions of visual and ego-vehicle motion information to prediction performance. Future work will explore the integration of language-based semantic information to further enhance pedestrian crossing intention prediction.